\documentclass[10pt,twocolumn,letterpaper]{article}

\usepackage[pagenumbers]{cvpr} % To force page numbers, e.g. for an arXiv version

\usepackage{graphicx}
\usepackage{amsmath}
\usepackage{amssymb}
\usepackage{booktabs}
\usepackage[dvipsnames]{xcolor}
\usepackage{multirow}
\usepackage{multicol}
\usepackage{sidecap}
\usepackage{stfloats}
\usepackage{float}

\usepackage[pagebackref,breaklinks,colorlinks]{hyperref}

\usepackage[capitalize]{cleveref}
\crefname{section}{Sec.}{Secs.}
\Crefname{section}{Section}{Sections}
\Crefname{table}{Table}{Tables}
\crefname{table}{Tab.}{Tabs.}

\def\cvprPaperID{****} % *** Enter the CVPR Paper ID here
\def\confName{CVPR}
\def\confYear{2023}

\begin{document}

\title{An Empirical Study on Clustering Pretrained Embeddings:\\Is Deep Strictly Better?}

\author{Tyler R.~Scott\\
Google Research\\
University of Colorado, Boulder\\
{\tt\small tylersco@google.com}
\and
Ting Liu\\
Google Research\\
{\tt\small liuti@google.com }
\and
Michael C.~Mozer\\
Google Research\\
{\tt\small mcmozer@google.com }
\and
Andrew C.~Gallagher\\
Google Research\\
{\tt\small agallagher@google.com }
}

\maketitle

\begin{abstract}
Recent research in clustering face embeddings has found that unsupervised, shallow, heuristic-based methods---including
$k$-means and hierarchical agglomerative clustering---underperform supervised, deep, inductive methods.
While the reported improvements are indeed impressive, experiments are mostly limited to face datasets, where the clustered embeddings
are highly discriminative or well-separated by class (Recall@1 above 90\% and often nearing ceiling), and the experimental methodology seemingly favors the deep methods.
We conduct a large-scale empirical study of 17 clustering methods across three datasets and obtain several robust findings. 
Notably, deep methods are surprisingly fragile for embeddings with more uncertainty, where they match or even perform worse than shallow, heuristic-based methods.
When embeddings are highly discriminative, deep methods do outperform the baselines, consistent with past results, but the margin between methods is much smaller than previously reported.
We believe our benchmarks broaden the scope of supervised clustering methods beyond the face domain and can serve as a foundation on which these methods could be improved.
To enable reproducibility, we include all necessary details in the appendices, and plan to release the code.
\end{abstract}

\section{Introduction}
\label{sec:intro}

Clustering---the process by which a set of items are semantically-partitioned into finite groups---
has been employed in many domains of computing including machine learning, computer graphics, information retrieval, and bioinformatics. It has numerous applications including interpretability, compression, visualization, outlier detector, and zero-shot classification.

While clustering methods have traditionally ingested raw features (\eg, pixels) as input, an alternative explored particularly in the face-verification literature is to use \emph{deep embeddings}. 
Deep embeddings are latent-feature vectors discovered by neural networks \cite{Guo2020,Lin2018,Nguyen2021,Otto2018,Shen2021,Shi2018,Wang2019Linkage,Yang2019,Yang2020,Ye2021,Zhan2018}. They form a natural space for clustering because they are learned with loss functions that encourage grouping of semantically-similar inputs (\eg, \cite{Boudiaf2020,Chopra2005,Deng2019,Hadsell2006,Snell2017,Weinberger2009,Sohn2016}) and explicit
representation of task-relevant features (\eg, \cite{Ridgeway2016,Schroff2015}).

Face verification systems rely on large backbone networks that produce highly-discriminative\footnote{The term \emph{discriminative} refers to the network's ability to separate embeddings that belong to the same identity or class from those that do not, as measured by a distance function.} embeddings of face images.
Improvements to these systems come largely through increasing the available labeled data, but with an expensive annotation cost.
To overcome the annotation challenge, recent work has successfully used clustering to pseudo-label \cite{He2018,Wang2019Linkage,Yang2019,Yang2020,Zhan2018,Shen2021,Nguyen2021}.
The approach is cyclical: train a face-embedding model, embed and cluster unsupervised face images, pseudo-label the images via the clustering assignments, merge the pseudo-labeled data into the training set, and re-train the model.
Since the crux of the approach is the clustering step, as improved clustering leads to better pseudo-labels and thus improved embedding quality, research has shifted from unsupervised, shallow, heuristic-based clustering methods \cite{Banerjee2005,Cheng1995,Ester1996,Frey2007,Gopal2014,Hornik2012,Jain2010,Kulis2012,Lin2018,Liu2014,Lloyd1982,Ng2001,Otto2018,Sculley2010,Shi2018,Sibson1973,Straub2015} to supervised, deep, inductive methods \cite{Guo2020,He2018,Nguyen2021,Shen2021,Wang2019Linkage,Yang2019,Yang2020,Ye2021,Zhan2018}, with claimed improvements upward of 20\% over $k$-means \cite{Lloyd1982} and hierarchical agglomerative clustering \cite{Sibson1973}, for example.

Even though improvements from the deep methods are impressive, the prior work has limitations.
First, by focusing primarily on face datasets, the clustering methods operate on embedding spaces that are highly discriminative and not representative of the many embedding spaces that could benefit from clustering.
Deng \etal \cite{Deng2019} show that both verification and Recall@1 \cite{Musgrave2020} performance of the embeddings are well-above 90\% and can even approach the ceiling across a range of face datasets, including MegaFace \cite{Ira2016}.
We also confirm that the embeddings for DeepFashion \cite{Liu2016deepfashion}---one of the only non-face datasets explored for clustering---exhibit Recall@1 above 90\%.
Second, the commonly-used face datasets (e.g., MS-Celeb1M \cite{Guo2016} and MegaFace \cite{Ira2016}) are no longer publicly available, with progress reliant on using unofficial, shared embeddings extracted from pretrained backbones.
Third, similar to Musgrave \etal \cite{Musgrave2020}, we find methodological choices that may implicitly favor the recent, deep methods; these choices include the lack of a validation split, monitoring test performance for hyperparameter tuning and early stopping, and copying baseline results from previous work.

For the reasons above, we conducted
a large-scale empirical study of methods for clustering pretrained embeddings.
We focused on outlining an end-to-end, reproducible pipeline, including dataset curation, backbone training, and clustering.
We then benchmarked 17 clustering methods---including two state-of-the-art, supervised, deep methods, GCN-VE \cite{Yang2020} and STAR-FC \cite{Shen2021}---on three datasets.
The first two datasets are Cars196 \cite{Krause2013} and Stanford Online Products (SOP) \cite{Song2016}.
We specifically chose to benchmark against Cars196 and SOP because they are: (1) popular in the embedding-learning literature \cite{Musgrave2020,Scott2021,Boudiaf2020}, (2) extend beyond the face-verification domain, and (3) produce embeddings that are significantly less discriminative, as discussed in \cref{sec:backbone_results}.
We investigate a third dataset that has similar statistics, similar Recall@1 performance, and thus similar clustering results compared to common face datasets (e.g., MS-Celeb1M) used solely in past research. The third dataset's purpose is simply to corroborate results from past research. To enable reproducibility, we include all necessary details for Cars196 and SOP in the appendices, and plan to release the code.

Our results indicate that the deep methods are indeed superior when operating on highly-discriminative embeddings, but their performance drops otherwise, matching or even underperforming the shallow, heuristic-based methods.
We conclude that the Recall@1 performance of the embeddings is generally an accurate proxy for predicting the benefits of deep clustering methods.
Additionally, where the deep methods do improve performance, the improvements are of much smaller magnitude than previously reported, likely as a result of our systematic training pipeline and matched hyperparameter tuning.
Lastly, we find that $\ell_2$-normalization of embeddings prior to clustering leads to a stark improvement across all heuristic-based methods, exploiting the geometrical properties of embeddings learned with softmax cross-entropy \cite{Scott2021}.

\section{Related Work}
\label{sec:related_work}

\subsection{Unsupervised Clustering}
\label{sec:unsupervised_clustering}

Clustering is normally characterized as an unsupervised learning task where methods leverage heuristics based on assumptions about the data-generation process or the structure of the data.
Adopting the terminology of Jain \cite{Jain2010}, we discuss four common classes of heuristics: partition-based, density-based, hierarchical, and graph-theoretic.

Methods using partition-based heuristics \cite{Frey2007,Hornik2012,Kulis2012,Lloyd1982,Sculley2010,Straub2015} have a representation of clusters and decide which data points belong to each cluster, in many cases, using a similarity or distance function.
The most popular method is $k$-means clustering \cite{Lloyd1982} which estimates empirical cluster centers minimizing within-cluster variance.
Other variants make assumptions about the geometry of the data, e.g., spherical $k$-means \cite{Hornik2012}, make assumptions about the scale of the data, e.g., mini-batch $k$-means \cite{Sculley2010}, or explore alternative formulations that remove the need for a predefined number of clusters, e.g., Dirichlet-process (DP) $k$-means \cite{Kulis2012,Straub2015}. 

Density-based methods assume that clusters represent high-density regions in the feature space and are separated from other clusters by low-density regions.
Popular density-based methods attempt to estimate cluster densities via expectation-maximization, e.g., the Gaussian mixture model (GMM) \& the von Mises--Fisher mixture model (vMF-MM) \cite{Banerjee2005}, variational inference \cite{Gopal2014}, Parzen windows, e.g., DBSCAN \cite{Ester1996} \& deep density clustering \cite{Lin2018}, and kernel density estimation, e.g., MeanShift \cite{Cheng1995}. 

Hierarchical methods form clusters by starting either with a singleton cluster containing all data points and iteratively splitting it (i.e., top-down or divisive) or with a cluster for each data point and iteratively merging them (i.e., bottom-up or agglomerative).
The split or merge is decided greedily using a linkage rule until a criteria is satisfied such as the total number of clusters.
We experiment with hierarchical agglomerative clustering (HAC) \cite{Sibson1973} and several popular linkage rules (single, complete, average, and Ward linkage), as well as approximate rank-order (ARO) \cite{Otto2018} based on the rank-order linkage metric \cite{Zhu2011}.

The final common class of heuristics use graph theoretics.
These methods represent data points as a graph with edges weighted by pairwise similarities.
One of the most popular methods, and the one we employ in our experiments, is spectral clustering \cite{Ng2001}. Spectral clustering performs $k$-means clustering on the eigenvectors of the normalized Laplacian matrix constructed from the graph's affinity matrix.

\subsection{Supervised Clustering}
\label{sec:supervised_clustering}

Recent research, instead of relying on heuristics, has proposed deep, supervised methods that inductively cluster.
These methods rely on supervision indicating which data points belong to the same cluster.
By \emph{inductive}, we refer to methods that learn a function, for example, a function that predicts if two data points should be clustered together, which can then be applied to unseen data.
This is in contrast to the unsupervised methods which lack generalizability from one dataset to another.
Generally, the idea is to leverage local and/or global structure in the feature space to learn what points belong to the same cluster or what pairs of points should be linked.
Clusters can then be formed using simple algorithms such as connected-component labeling.

Consensus-driven propagation (CDP) \cite{Zhan2018} uses an ensemble of backbone networks to produce statistics across many $k$-nearest-neighbor affinity graphs and then trains a \emph{mediator} network to predict links between data points from which clusters are formed.
Wang \etal \cite{Wang2019Linkage} improves link prediction by capturing structure in local affinity graphs directly with graph convolutional networks (GCNs) \cite{Kipf2017}.
A series of GCN-methods followed that incorporate both local and global structure via density information \cite{Guo2020} and multi-scale views \cite{Yang2019}, for example.
Alternatives beyond GCNs have been proposed such as using self-attention via transformers \cite{Nguyen2021,Ye2021} and learning an agglomerative clustering policy with inverse reinforcement learning \cite{He2018}.

Our experiments include results from CDP, as well as two state-of-the-art GCN approaches: GCN-VE \cite{Yang2020} and STAR-FC \cite{Shen2021}.
GCN-VE is a supervised approach that trains two GCNs.
The first, GCN-V, estimates a confidence for each data point based on a supervised density measure.
The second, GCN-E, constructs subgraphs based on the estimated confidences and predicts pairwise links.
The links are used in a \emph{tree-deduction} algorithm to construct final clusters.
STAR-FC trains a single GCN to predict pairwise links, but uses a structure-preserving sampling strategy to train the GCN with both local and global graph structure.
During inference, links are predicted followed by \emph{graph parsing and refinement} steps to construct clusters.

Consistent with past work, we do not train an ensemble and mediator for CDP, but rather use just the unsupervised \emph{label-propagation} step to form clusters.
Note that this removes all supervised components of CDP, which is why we classify it as an unsupervised method in \cref{sec:results}.

\section{Methodology}
\label{sec:methodology}

Our pipeline for clustering pretrained embeddings involves three steps: dataset curation, backbone training, and clustering.
We discuss the methodology for each of the steps below.

\subsection{Dataset Curation}
\label{sec:dataset_curation}

\begin{table}[t]
\begin{center}
\small
\begin{tabular}{@{}ccccccc@{}}
\toprule
& \multicolumn{3}{c}{Cars196} & \multicolumn{3}{c}{SOP} \\
\cmidrule(lr){2-4}
\cmidrule(lr){5-7}
& $m$ & $n$ & $\frac{m}{n}$ & $m$ & $n$ & $\frac{m}{n}$ \\
\midrule
Backbone Train & $3,963$ & $49$ & $81$ & $32,277$ & $5,658$ & $6$ \\
Clustering Train & $4,091$ & $49$ & $83$ & $27,265$ & $5,658$ & $5$ \\
Validation & $4,058$ & $49$ & $83$ & $32,734$ & $5,658$ & $6$ \\
Test & $4,073$ & $49$ & $83$ & $27,777$ & $5,660$ & $5$ \\
\bottomrule
\end{tabular}
\end{center}

\caption{Dataset splits with number of instances, $m$, number of classes, $n$, and approximate number of instances per class, $\frac{m}{n}$, for Cars196 and SOP. The classes in each split of are disjoint.}
\label{tab:dataset_stats}
\end{table}

We consider three datasets for our experiments: Cars196 \cite{Krause2013}, Stanford Online Products (SOP) \cite{Song2016}, and a private, third dataset, \emph{Dataset 3}.
Each dataset is partitioned into a \emph{backbone train} split, used to train the backbone, a \emph{clustering train} split, used only to train the deep clustering methods, a \emph{validation} split, used for hyperparameter tuning and early stopping for the backbone and deep clustering methods, and one or more \emph{test} splits, used to fit the unsupervised clustering methods and evaluate all clustering methods.

Cars196 and SOP have a single test split while Dataset 3 has five test splits that cumulatively grow in size, enabling investigation into the scalability of clustering methods.
All splits, except for the test splits of Dataset 3, have disjoint sets of classes.
\Cref{tab:dataset_stats} has per-split statistics on the number of instances, number of classes, and approximate number of instances per class for both Cars196 and SOP.
Dataset 3 has $\mathcal{O}(100,000)$ instances, $\mathcal{O}(1,000)$ classes, and roughly $\mathcal{O}(100)$ instances per class for all splits.

We acknowledge that the deep methods are provided an advantage through access to the train-clustering split, which is unavailable to the shallow methods.
Ideally, the shallow methods would make use of the additional split of data, however, we chose to match the protocol of past research on clustering face embeddings.

\subsection{Backbone Training}
\label{sec:backbone_training}

We use a ResNet50 \cite{He2016} for the backbone with an additional fully-connected layer added after global average pooling and prior to the classification head, which produces a 256D embedding.
We chose a ResNet50 backbone to match prior research on face clustering \cite{Nguyen2021, Shen2021, Yang2019, Yang2020, Ye2021}, but note that consistent clustering results were found with an Inception v3 backbone \cite{He2018}.
The model is trained with cosine softmax cross-entropy---based on findings from Scott \etal \cite{Scott2021}---using the backbone-train split of each dataset, with Recall@1 monitored on the validation split for early stopping and hyperparameter tuning.
A hyperparameter search was performed for each of the datasets.
Additional details on training the backbone, including specifics on the architecture, data augmentation, and hyperparameters are included in \cref{sec:backbone_details}.

\subsection{Clustering}
\label{sec:clustering}

\begin{figure*}[t]
\centering
\includegraphics[width=1.0\textwidth]{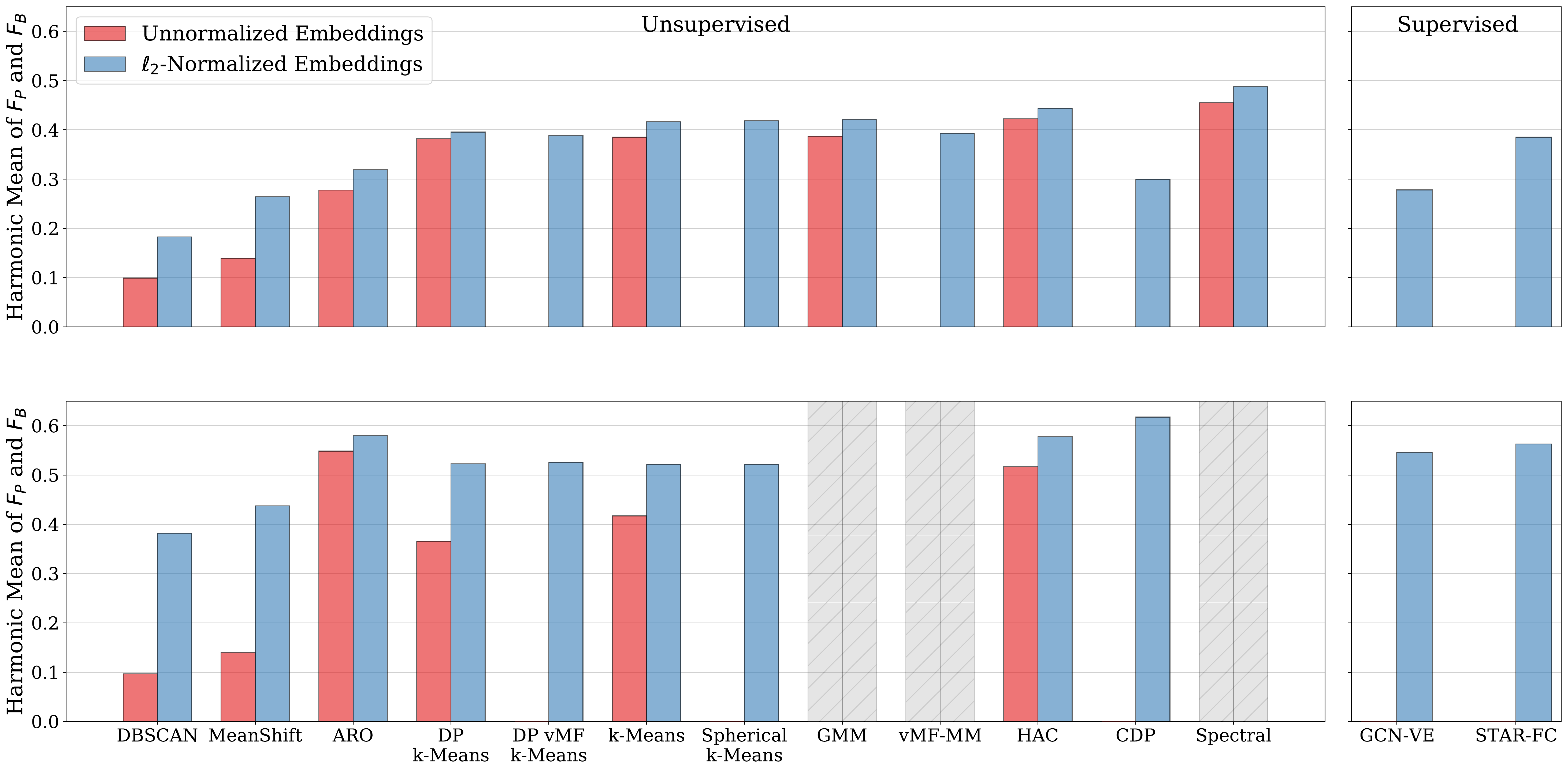}
\caption{Harmonic mean of Pairwise ($F_P$) and BCubed ($F_B$) F-scores across clustering methods for Cars196 (top) and SOP (bottom). The left and right pane contain results for unsupervised and supervised clustering methods, respectively, and the red and blue bars indicate clustering of unnormalized and $\ell_2$-normalized embeddings, respectively. GMM, vMF-MM, and Spectral could not be run on SOP due to runtime inefficiencies, indicated by the gray, hatched bars. The methods that omit a result for unnormalized embeddings assume $\ell_2$-normalized embeddings, by default.}
\label{fig:cars_sop_cluster}
\end{figure*}

The clustering methodology varies between the supervised, deep methods (\eg, GCN-VE and STAR-FC) and the unsupervised, shallow methods (\eg, $k$-means).
We discuss each below.
\Cref{sec:clustering_details} contains additional methodological details, architecture details for GCN-VE and STAR-FC, and an enumeration of all hyperparameters and their tuned values across all clustering methods.

For supervised clustering, we compute the 256D embeddings for the clustering-train, validation, and test splits.
Methods are trained using the clustering-train embeddings from each dataset with the loss monitored on the validation split for early stopping and hyperparameter tuning.
A hyperparameter search was performed for each of the datasets and we report results using the set of hyperparameter values associated with maximal clustering performance on the validation split.
The model is then applied, inductively, on the test embeddings and final performance is reported.

\begin{table}[b]
\begin{center}
\small
\scalebox{0.93}{
\begin{tabular}{@{}cccc@{}}
\toprule
\multicolumn{4}{c}{Unnormalized Embeddings} \\
\midrule
& Clustering & \multirow{2}{*}{Validation} & \multirow{2}{*}{Test} \\
& Train & & \\
\midrule
Cars196 & $0.67$ & $0.76$ & $0.76$ \\
SOP & $0.72$ & $0.70$ & $0.73$ \\
Dataset 3 & $0.91$ & $0.91$ & $0.91$, $0.89$, $0.88$, $0.87$, $0.86$ \\
\bottomrule
\end{tabular}}
\end{center}

\begin{center}
\small
\scalebox{0.93}{
\begin{tabular}{@{}cccc@{}}
\toprule
\multicolumn{4}{c}{$\ell_2$-Normalized Embeddings} \\
\midrule
& Clustering & \multirow{2}{*}{Validation} & \multirow{2}{*}{Test} \\
& Train & & \\
\midrule
Cars196 & $0.69$ & $0.80$ & $0.79$ \\
SOP & $0.75$ & $0.74$ & $0.77$ \\
Dataset 3 & $0.94$ & $0.95$ & $0.94$, $0.93$, $0.92$, $0.91$, $0.91$ \\
\bottomrule
\end{tabular}}
\end{center}

\caption{Recall@1 on the clustering-train, validation, and test splits of each dataset for both unnormalized and $\ell_2$-normalized embeddings. The comma-separated values for the test column of Dataset 3 represent the five test splits.}
\label{tab:backbone_recall}
\end{table}

In contrast, the unsupervised methods operate directly on the embeddings from the test split(s).
A hyperparameter search was also conducted for the unsupervised methods, however, we report results using the set of values associated with maximal clustering performance on the test split(s).
We admit that optimizing for test performance appears to provide an unfair advantage to the unsupervised methods, however, the supervised methods: (1) have access to a split of data unavailable to the unsupervised methods, (2) have thousands of learnable parameters, and (3) have 12 and 16 hyperparameters for STAR-FC and GCN-VE, respectively, compared to at-most 4 hyperparameters for the unsupervised methods.

There are no alterations to the embeddings other than an $\ell_2$-normalization step that we found to unequivocally improve clustering performance.
\Cref{sec:clustering_results} further discusses the benefits of $\ell_2$-normalization.

\section{Experimental Results}
\label{sec:results}

\begin{table*}[t]
\begin{center}
\small
\begin{tabular}{@{}ccccccccccc@{}}
\toprule
& \multicolumn{2}{c}{Test \#1} & \multicolumn{2}{c}{Test \#2} & \multicolumn{2}{c}{Test \#3} & \multicolumn{2}{c}{Test \#4} & \multicolumn{2}{c}{Test \#5} \\
\midrule
& $F_P$ & $F_B$ & $F_P$ & $F_B$ & $F_P$ & $F_B$ & $F_P$ & $F_B$ & $F_P$ & $F_B$ \\
\midrule
Minibatch $k$-Means & $0.64$ & $0.66$ & $0.60$ & $0.61$ & $0.57$ & $0.58$ & $0.55$ & $0.56$ & $0.54$ & $0.55$ \\
DP $k$-Means & $0.67$ & $0.68$ & $0.62$ & $0.60$ & $0.61$ & $0.59$ & \color{blue}$0.45$ & \color{blue}$0.53$ & \color{blue}$0.43$ & \color{blue}$0.49$ \\
DP vMF $k$-Means & $0.72$ & $0.70$ & $0.67$ & $0.65$ & $0.60$ & $0.58$ & $0.61$ & $0.59$ & $0.61$ & $0.59$ \\
HAC & $0.74$ & $0.75$ & $0.69$ & $0.72$ & $0.66$ & $0.69$ & \multicolumn{2}{c}{DNC} & \multicolumn{2}{c}{DNC} \\
ARO & $0.55$ & $0.59$ & $0.53$ & $0.59$ & $0.50$ & $0.56$ & $0.50$ & $0.51$ & $0.48$ & $0.49$ \\
DBSCAN & $0.32$ & $0.41$ & $0.21$ & $0.40$ & $0.20$ & $0.20$ & $0.20$ & $0.20$ & $0.19$ & $0.20$ \\
CDP & $0.62$ & $0.62$ & $0.59$ & $0.59$ & $0.56$ & $0.56$ & $0.54$ & $0.54$ & $0.52$ & $0.53$ \\
\midrule
GCN-VE & $\boldsymbol{0.78}$ & $\boldsymbol{0.79}$ & $\boldsymbol{0.74}$ & $\boldsymbol{0.75}$ & $\boldsymbol{0.71}$ & $\boldsymbol{0.72}$ & $\boldsymbol{0.68}$ & $\boldsymbol{0.70}$ & $0.62$ & $\boldsymbol{0.65}$ \\
STAR-FC & $0.65$ & $0.75$ & $0.64$ & $0.72$ & $0.65$ & $0.69$ & $0.65$ & $0.66$ & $\boldsymbol{0.63}$ & $0.64$ \\
\bottomrule
\end{tabular}
\end{center}

\begin{center}
\small
\begin{tabular}{@{}cccccc@{}}
\toprule
& Test \#1 & Test \#2 & Test \#3 & Test \#4 & Test \#5 \\
\midrule
Minibatch $k$-Means & $0.2$h & $0.7$h & $1.3$h & $2.5$h & $4$h \\
DP $k$-Means & $13$h & $70$h & $127$h & \color{blue}$253$h & \color{blue}$324$h \\
DP vMF $k$-Means & $29$h & $98$h & $27$h & $9$h & $12$h \\
HAC & $6$h & $35$h & $97$h & DNC & DNC \\
ARO & $0.5$h & $1.8$h & $4$h & $3$h & $9$h \\
DBSCAN & $0.1$h & $0.4$h & $0.8$h & $1.4$h & $1.9$h \\
CDP & $0.4$h & $1.3$h & $3$h & $5$h & $8$h \\
\midrule
GCN-VE & $132$h, $0.3$h & $132$h, $0.7$h & $132$h, $1.6$h & $132$h, $2.8$h & $37$h, $2.5$h \\
STAR-FC & $45$h, $0.2$h & $45$h, $0.6$h & $45$h, $0.9$h & $45$h, $1.3$h & $45$h, $1.8$h \\
\bottomrule
\end{tabular}
\end{center}

\caption{Pairwise ($F_P$) and BCubed ($F_B$) F-scores (top) and time to cluster in hours (bottom) across clustering methods for all test splits of Dataset 3. \emph{DNC} indicates that the method \emph{did not converge} in a reasonable amount of time. The time to cluster for GCN-VE and STAR-FC contains the train time and test time separated by a comma. Boldface indicates the highest value for a metric. \color{blue}Blue \color{black} indicates that the particular run used unnormalized embeddings instead of $\ell_2$-normalized embeddings.}
\label{tab:dataset3_cluster}
\end{table*}

Using the methodology outlined in \cref{sec:methodology}, we conduct extensive experiments on Cars196, Stanford Online Products (SOP), and Dataset 3 across 17 clustering methods.
We begin by measuring the discriminability of the embeddings learned by the backbones and then move on to the clustering results.
Details on compute resources are included in \cref{sec:compute_resources}.

\subsection{Backbone Results}
\label{sec:backbone_results}

One connection we attempt to make in our experimentation is between embedding discriminability and clustering performance: the likelihood of deep clustering methods outperforming shallow methods increases as classes are better discriminated.
Thus, we quantify embedding discriminability across all three datasets for reference in coming sections.
\Cref{tab:backbone_recall} contains Recall@1 for the clustering-train, validation, and test splits of each dataset.
The top table uses unnormalized embeddings (\ie, embeddings that lie in Euclidean space) and the bottom table uses $\ell_2$-normalized embeddings (\ie, embeddings projected onto the surface of a unit hypersphere).
Consistent with Scott \etal \cite{Scott2021}, we find that $\ell_2$-normalization leads to strictly improved discrimination, as it removes intra-class variance.

We chose Cars196, SOP, and Dataset 3 as benchmarks because the discriminability of the embeddings varies significantly, with Cars196 and SOP performance between 70\% and 80\%, and Dataset 3 above 90\%, for $\ell_2$-normalized embeddings.
We emphasize that this is a confound not considered in past work on face clustering.
The datasets previously explored all have Recall@1 performance above 90\% and some even near ceiling \cite{Deng2019}.

\subsection{Clustering Results}
\label{sec:clustering_results}

We measure clustering performance on the test splits of all datasets.
The clustering results focus on two metrics: Pairwise ($F_P$) \cite{Shi2018} and BCubed ($F_B$) \cite{Amigo2009} F-scores.
Because the metrics have different emphases---specifically, $F_P$ emphasizes fidelity of larger clusters and $F_B$ emphasizes fidelity of clusters proportional to their size---we report their harmonic mean when not reporting both, individually.

\Cref{fig:cars_sop_cluster} shows clustering performance for Cars196 (top) and SOP (bottom) across 12 unsupervised methods and the 2 supervised methods.
\Cref{sec:cars_sop_clustering_results} contains tabulated results used to construct the figure, as well as additional metrics such as normalized mutual information.
The red and blue bars represent performance on unnormalized and $\ell_2$-normalized embeddings, respectively.
There are two key takeaways: (1) using $\ell_2$-normalized embeddings consistently improves clustering performance, sometimes upward of 30\%, and (2) the supervised, deep methods are outperformed by unsupervised, shallow methods and are not even among the top-3 performers, contrary to past work reporting consistent state-of-the-art performance.
Among the unsupervised methods, there is no clear best-performing method, however, HAC performs strongly on both datasets.
As a point of comparison, HAC outperforms GCN-VE and STAR-FC by 16\% and 6\% for Cars196, respectively, and 3\% and 1\% for SOP, respectively.
The results on Cars196 and SOP highlight a downside of the deep methods, particularly that they become fragile in the presence of uncertainty (\ie, less discriminability) in the embedding space.

To verify we can reproduce the benefits of GCN-VE and STAR-FC, we performed a similar clustering analysis on Dataset 3, where Recall@1 is above 90\% for all splits.
\Cref{tab:dataset3_cluster} contains Pairwise and BCubed F-scores (top) and the time to cluster (bottom) across the five test splits.
Notice that the results are inverted: GCN-VE and STAR-FC consistently outperform the shallow methods.
However, the margin between HAC, for example, and GCN-VE is less than 5\%---for the test splits where HAC converged.
Comparing to the results in Yang \etal \cite{Yang2020} and Shen \etal \cite{Shen2021}, GCN-VE is shown to consistently outperform HAC by upward of 20\%.
In agreement with results reported above for Cars196 and SOP, $\ell_2$-normalized embeddings, again, outperform unnormalized embeddings, compatible with prior work on embedding-learning and face verification \cite{Scott2021,Deng2019,Liu2017,Ranjan2017,Ranjan2019,Wang2017normface,Wang2018,Wang2018a}.
The only exception is for test splits \#4 and \#5 for DP $k$-means, where unnormalized embeddings were superior.

In addition to state-of-the-art performance, another benefit underlined in past work for GCN-VE and STAR-FC is their efficient inference time.
We computed the time to cluster the Dataset 3 test splits in the bottom of \cref{tab:dataset3_cluster}.
We also find that GCN-VE and STAR-FC are among the most efficient methods at inference time, outpaced only by DBSCAN and minibatch $k$-means.
However, one factor not presented in past work is the \emph{total} time to employ deep clustering methods, that is, including the additional time to train the method.
We find that training can take tens of hours and when factored in with the inference time, presents a trade-off for the deep methods.
Depending on the application, GCN-VE and STAR-FC may only be used for inference one time, whereas if they are used in a production system may be used for inference hundreds of times.
When considering the total time, in addition to their marginal improvements over much simpler methods, we encourage inspection into the amortized time (\ie, the dispersion of training time amongst the expected number of inference runs).
In the case of fewer inference runs, methods such as minibatch $k$-means may provide adequate performance while producing a more-efficient amortized runtime.

\subsubsection{Investigation into GCN-VE}
\label{sec:gcn_ve_investigation}

\begin{table}[t]
\begin{center}
\small
\begin{tabular}{@{}ccccc@{}}
\toprule
& \multicolumn{2}{c}{Cars196} & \multicolumn{2}{c}{SOP} \\
\cmidrule(lr){2-3}
\cmidrule(lr){4-5}
& $F_P$ & $F_B$ & $F_P$ & $F_B$ \\
\midrule
GCN-VE \cite{Yang2020} & $0.26$ & $0.37$ & $0.44$ & $0.63$ \\
GCN-VE (ours) & $0.22$ & $0.37$ & $0.47$ & $0.65$ \\
\bottomrule
\end{tabular}
\end{center}
\caption{Comparison of Pairwise ($F_P$) and BCubed ($F_B$) F-scores for the open-source GCN-VE code and our reimplementation on Cars196 and SOP.}
\label{tab:gcn_ve_verify}
\end{table}

\begin{table}[t]
\begin{center}
\small
\scalebox{0.96}{
\begin{tabular}{@{}ccccccc@{}}
\toprule
& \multicolumn{2}{c}{\multirow{2}{*}{Cars196}} & \multicolumn{2}{c}{\multirow{2}{*}{SOP}} & \multicolumn{2}{c}{Dataset 3} \\
& & & & & \multicolumn{2}{c}{Test \#1} \\
\cmidrule(lr){2-3}
\cmidrule(lr){4-5}
\cmidrule(lr){6-7}
& $F_P$ & $F_B$ & $F_P$ & $F_B$ & $F_P$ & $F_B$ \\
\midrule
Tree Deduction & $0.13$ & $0.36$ & $0.44$ & $0.63$ & $0.38$ & $0.41$ \\
+ $\ell_2$-Norm & $\boldsymbol{0.25}$ & $\boldsymbol{0.37}$ & $0.44$ & $0.63$ & $0.62$ & $0.65$ \\
+ GCN-E & $0.19$ & $\boldsymbol{0.37}$ & $0.44$ & $0.62$ & $0.62$ & $0.65$ \\
\midrule
GCN-VE & $0.22$ & $\boldsymbol{0.37}$ & $\boldsymbol{0.47}$ & $\boldsymbol{0.65}$ & $\boldsymbol{0.78}$ & $\boldsymbol{0.79}$ \\
\bottomrule
\end{tabular}}
\end{center}
\caption{Ablation study comparing GCN-VE to simpler variants. The reported results are Pairwise ($F_P$) and BCubed ($F_B$) F-scores on Cars196, SOP, and test split \#1 of Dataset 3. Boldface indicates the highest value for a metric.}
\label{tab:gcn_ve_variants}
\end{table}

Due to the unexpected results of GCN-VE on Cars196 and SOP, we conducted further investigation of the method to verify our findings.
First, we compared our reimplementation of GCN-VE to the open-source version.
\Cref{tab:gcn_ve_verify} compares the two implementations on Cars196 and SOP.
We confirm that our results, within reasonable variance, match the implementation from Yang \etal \cite{Yang2020}.
Next, we ablated the components of GCN-VE to better understand which contributed to final performance.
In general, GCN-VE works by: (1) estimating a confidence for each embedding via the GCN-V graph convolutional network, (2) constructing subgraphs using the estimated confidences, (3) estimating which edges in the subgraphs are valid connections via the GCN-E graph convolutional network, and (4) running a tree-deduction algorithm to form the final clusters.

We consider three ablated variants of GCN-VE, described below.
All that is needed to form clusters is to run tree deduction on a $k$-nearest-neighbors subgraph with edges pruned by a distance threshold.
We refer to this first ablation as \emph{Tree Deduction}.
Based on empirical analysis from Scott \etal \cite{Scott2021}, the embedding $\ell_2$-norm is a measure of confidence, thus, we can replace the GCN-V network with the $\ell_2$-norm of each embedding.
Additionally, we can replace the GCN-E network simply by considering edges valid if they are connected to higher-confidence embeddings within a distance threshold, and run tree deduction as is.
We refer to this second ablation as \emph{Tree Deduction + $\ell_2$-norm}.
The third ablation, \emph{Tree Deduction + $\ell_2$-norm + GCN-E}, reintroduces GCN-E, but still uses $\ell_2$-norm instead of the GCN-V network.

\Cref{tab:gcn_ve_variants} measures clustering performance of the three ablations against GCN-VE for Cars196, SOP, and test split \#1 of Dataset 3.
Interestingly, one can recover the majority of GCN-VE performance with the Tree Deduction + $\ell_2$-norm variant, corroborating results from Scott \etal \cite{Scott2021} on $\ell_2$-norm conveying embedding confidence.
Additionally, the critical component of GCN-VE is not the GCN-E network, but GCN-V.
Tree Deduction + $\ell_2$-norm performs as well, if not better, than Tree Deduction + $\ell_2$-norm + GCN-E.

\subsubsection{Degradation Study}
\label{sec:degradation_study}

\begin{table}[t]
\begin{center}
\small
\begin{tabular}{@{}ccccc@{}}
\toprule
& \multirow{2}{*}{32D} & \multirow{2}{*}{64D} & \multirow{2}{*}{128D} & 256D \\
& & & & (original) \\
\midrule
Dataset 3 & \multirow{2}{*}{$0.68$} & \multirow{2}{*}{$0.86$} & \multirow{2}{*}{$0.92$} & \multirow{2}{*}{$0.94$} \\
Clustering Train & & & & \\
\bottomrule
\end{tabular}
\end{center}
\caption{Recall@1 on the clustering-train split of Dataset 3 as the embedding dimensionality increases.}
\label{tab:dataset3_degradation_recall}
\end{table}

\begin{figure}[t]
\centering
\includegraphics[width=0.48\textwidth]{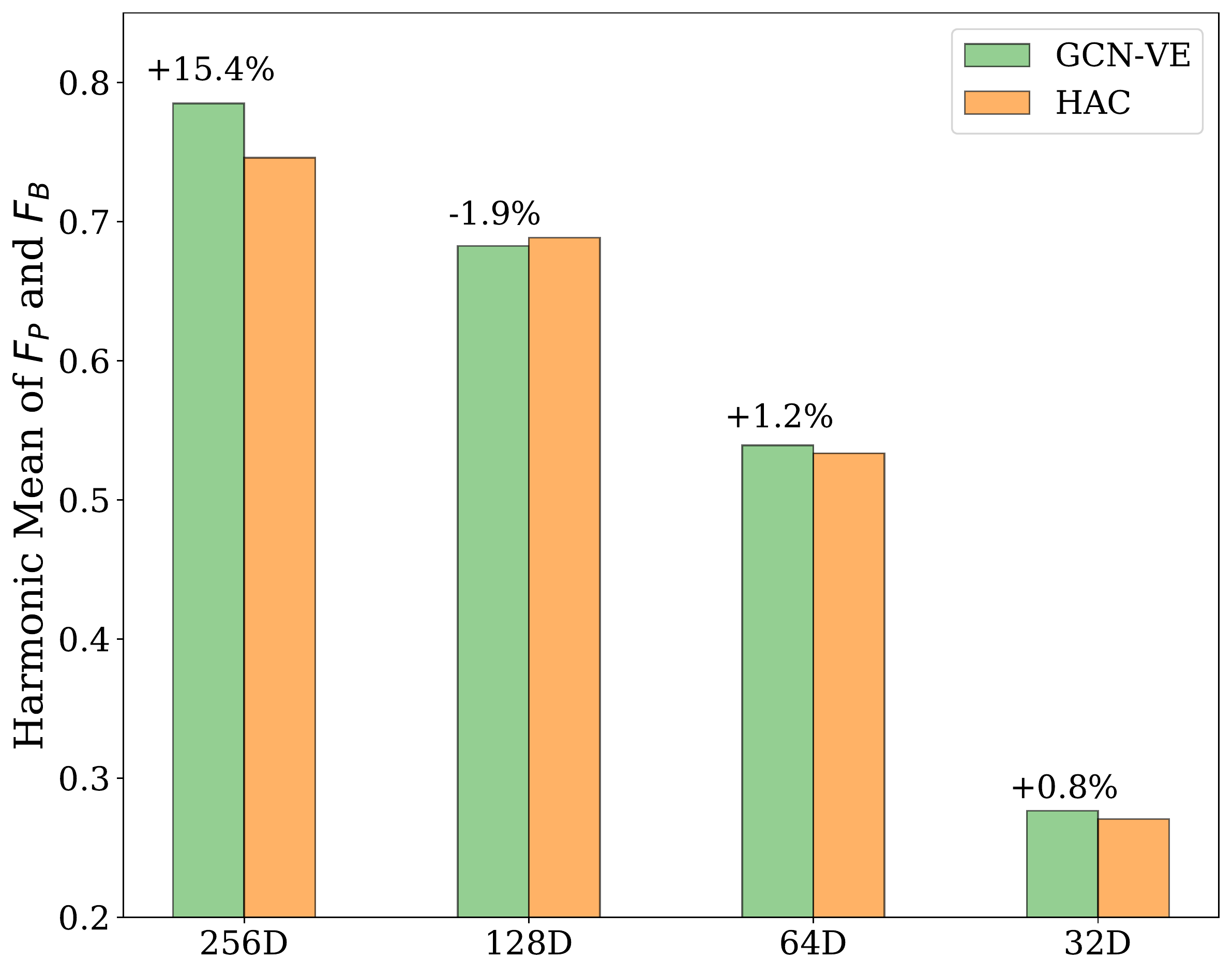}
\caption{Harmonic mean between Pairwise ($F_P$) and BCubed ($F_B$) F-scores of GCN-VE and HAC on test split \#1 of Dataset 3 for decreasing embedding dimensionalities. The percentages above the GCN-VE (green) bars indicate its relative reduction in error compared to HAC.}
\label{fig:dataset3_degradation}
\end{figure}

While the experimental results thus far support our claim that the benefits of deep clustering methods are tied to embedding discriminability, we admit there are confounds.
First, Cars196 and SOP are non-face datasets and second, they have very different dataset statistics compared to Dataset 3 and the historically-used face datasets, summarized in \cref{tab:dataset_stats}.
To remove these confounds, we perform a degradation study on Dataset 3 where we randomly select a subset of embedding dimensions to ignore, thus decreasing the discriminability.
We remove the same embedding dimensions for all splits of the data to ensure they are compatible.
\Cref{tab:dataset3_degradation_recall} measures the Recall@1 performance of the clustering-train split of Dataset 3 for varying embedding dimensionalities.
As expected, the lower-dimensional spaces have reduced discriminability, for example, the 32D space has Recall@1 dropping 26\% compared to the original 256D embedding.
\Cref{fig:dataset3_degradation} measures the harmonic mean of $F_P$ and $F_B$ for GCN-VE and HAC on test split \#1 of Dataset 3 as we decrease the embedding dimensionality.
The tabulated results composing the figure are included in \cref{sec:dataset3_degradation}.
The percentage above the GCN-VE (green) bars indicates the relative error reduction that GCN-VE achieves over HAC.
Note that for the original 256D embedding space, GCN-VE has a 15\% relative error reduction over HAC, but by projecting the embedding space to 128D---decreasing embedding discriminability by a mere 2\%---GCN-VE loses all advantage over HAC.
Thus, while we do not see as large of a performance drop for GCN-VE relative to HAC, comparatively, as we do for Cars196, the degradation study is consistent with our SOP results.

\section{Discussion and Conclusions}
\label{sec:discussion}

Many of the innovations in supervised methods for clustering pretrained embeddings are tied to face verification.
The methods were benchmarked primarily on face datasets and motivated by enlarging them via pseudo-labeling the vast amounts of unlabeled face data.
Common face datasets (\eg, MS-Celeb1M \cite{Guo2016} and MegaFace \cite{Ira2016}) are no longer publicly available, with progress reliant on using unofficial, shared embeddings extracted from pretrained backbones.
Additionally, the impressive improvements of supervised methods are bound to face embeddings, shown to have verification and Recall@1 performance well above 90\% and in some cases near ceiling \cite{Deng2019}.
Finally, results of recent work may implicitly favor the novel, supervised methods based on methodological choices such as the lack of a validation split and hyperparameter tuning based on test performance.

Our goals in conducting a large-scale empirical study on clustering pretrained embeddings are to: (1) broaden the scope of supervised clustering methods beyond faces, (2) present an end-to-end pipeline including backbone training along with results for benchmarking future methods, and (3) benchmark the robustness of supervised, deep methods on embeddings with less discriminability. 
We do so by presenting benchmarks for 17 clustering methods on three datasets: Cars196, SOP, and a third dataset, Dataset 3.
Cars196 and SOP are not only from diverse visual domains, but have very different dataset statistics (\ie, \cref{tab:dataset_stats}) and embedding discriminability (\ie, \cref{tab:backbone_recall}) compared to the historically-used face datasets.
We emphasize that Dataset 3 was chosen to corroborate results from past embedding-clustering research.
We discuss conclusions from our study below.

\paragraph{Embedding discriminability vs.~supervised clustering performance.} The main hypothesis we verify is the existence of a relationship between how discriminative the embeddings are (\ie, Recall@1) and the benefits of supervised, deep clustering methods.
We see for Cars196 and SOP---where embedding Recall@1 is between 70\% and 80\%---that the state-of-the-art supervised methods are not among the top 3 performers.
They are outperformed by HAC and spectral clustering, for example.
In contrast, for Dataset 3---where embedding Recall@1 is above 90\%---GCN-VE is a consistent state-of-the-art method followed by STAR-FC.
To remove any confounds caused by the domain or the dataset statistics, we show that we can remove all benefits of the supervised methods by randomly projecting the Dataset 3 embeddings from 256D to 128D (\ie, decreasing Recall@1 by a mere 2\%).
We leave it to future work to amend the fragility of supervised, deep methods, but we hypothesize that it may be caused by susceptibility of these methods to overfit spurious correlations in the embedding space that aren't generalizable and that simpler heuristics are more effective in cases of high embedding uncertainty.

\paragraph{Consistency with results from past work.} We believe certain methodological choices such as not using a validation split and using the test split for hyperparameter tuning and early stopping may have implicitly favored the deep methods, which are more flexible and have significantly more hyperparameters than the shallow methods.
Instead, we use a validation set for hyperparameter selection and early stopping for the deep methods, as well as reimplement and evaluate all baselines, and find that the unsupervised methods are actually competitive.
For example, HAC only underperforms state-of-the-art on Dataset 3 by 5\%, whereas past work indicates it can underperform by 20\% or more consistently \cite{Yang2020,Shen2021}.
While supervised methods are clearly superior on Dataset 3, the margin of improvement is of much smaller magnitude than previously reported.

\paragraph{Benefits of $\ell_2$-normalization.} It has been shown that softmax cross-entropy and its variants discover the most-discriminative embedding spaces \cite{Scott2021,Musgrave2020,Boudiaf2020,Tian2020}.
Due to the structure of the embeddings, the majority of intra-class variance extends outward from the origin.
As a result, Scott \etal \cite{Scott2021} verify that $\ell_2$-normalization of the embeddings leads to robust Recall@1 improvements, which we find transfers directly to clustering performance, as well (\ie, \cref{fig:cars_sop_cluster}).
For clustering pretrained embeddings, particularly trained with some form of softmax cross-entropy, we recommend $\ell_2$-normalization prior to clustering.

\paragraph{Performance vs.~runtime tradeoff.} One benefit of the supervised methods is their efficient inference time.
GCN-VE and STAR-FC are outpaced only by DBSCAN and minibatch $k$-means in \cref{tab:dataset3_cluster} for Dataset 3.
However, the time to train supervised methods is non-negligible and one should consider amortized runtime when choosing among methods.
If the goal is to use clustering for visualization, for example, and inference is only going to run once, the total runtime may be too costly given the marginal improvements over much simpler and faster unsupervised methods.

\paragraph{On the value of shallow methods.} Given recent trends, one might expect the supervised, deep methods to be strictly superior to the unsupervised, shallow methods that have become commonplace for clustering.
However, broadening the scope beyond the face domain has underlined their fragility in the presence of embedding uncertainty and emphasized the value of shallow methods.
We find that fundamental methods such as spectral clustering and HAC can outperform GCN-VE and STAR-FC despite having three times fewer hyperparameters and no learnable parameters.
By proposing new benchmarks on Cars196 and SOP, we hope that our empirical study serves as the foundation for which supervised methods can be further improved.

\section{Ethical Considerations}
\label{sec:ethical_considerations}

A major goal of our research is to broaden the study of clustering pretrained embeddings beyond the facial verification domain, which we do by providing benchmarks on Cars196 and SOP.
We appreciate the sensitivity associated with research on human data, including images of faces, and emphasize that clustering should only be employed in contexts that are responsible, socially beneficial, and fair, for example, in personalization of products for better user experiences, and not as part of harmful technologies or systems.
Additionally, we acknowledge that non-human datasets, including Cars196 and SOP, can exhibit bias through lack of representation.
We chose to report only on the downstream clustering performance, as to not reflect on or reinforce biases present in the data.

% \clearpage

{\small
\bibliographystyle{ieee_fullname}
\bibliography{main}
}

\clearpage

\onecolumn{
\appendix
\section{Compute Resources}
\label{sec:compute_resources}

Each backbone was trained using a single NVIDIA Tesla T4, 16 CPUs, and 64 GB of RAM on Google Cloud Platform.
Code was implemented using PyTorch v1.6.0 and Python 3.8.2 on Ubuntu 18.04.

For clustering, all methods had access to 16 CPUs, 96 GB of RAM, and a single NVIDIA Tesla T4, when appropriate.
Code was implemented with PyTorch v1.6.0, scikit-learn v0.24.2, FAISS v1.7.1, and Python 3.8.2 on Ubuntu 18.04.
Additionally, we note that both HAC and spectral clustering could be made faster by computing the nearest-neighbor subgraph on GPU instead of CPU.

\section{Cars196 and SOP Tabulated Results}
\label{sec:cars_sop_clustering_results}

The tables below contains clustering results for Cars196 and SOP.
The $F_P$ and $F_B$ values are used to compute the harmonic means for $\ell_2$-normalized embeddings presented in \cref{fig:cars_sop_cluster}.

\begin{table}[H]
\begin{center}
\scalebox{0.96}{
\begin{tabular}{@{}cccccccc@{}}
\toprule
\multicolumn{8}{c}{Cars196} \\
\midrule
& Adj. Rand Index & NMI & AMI & $F_P$ & $F_B$ & Time to Cluster (s) & Number of Clusters \\
\midrule
DBSCAN & $0.10$ & $0.63$ & $0.28$ & $0.12$ & $0.35$ & $3$ & $2,425$ \\
MeanShift & $0.18$ & $0.65$ & $0.42$ & $0.20$ & $0.38$ & $177$ & $1,407$ \\
ARO & $0.28$ & $0.69$ & $0.49$ & $0.29$ & $0.36$ & $3$ & $1,415$ \\
DP $k$-Means & $0.37$ & $0.69$ & $0.61$ & $0.38$ & $0.41$ & $10$ & $403$ \\
DP vMF $k$-Means & $0.36$ & $0.67$ & $0.61$ & $0.38$ & $0.40$ & $12$ & $155$ \\
$k$-Means & $0.40$ & $0.65$ & $0.62$ & $0.41$ & $0.42$ & $12$ & $50$ \\
Spherical $k$-Means & $0.40$ & $0.65$ & $0.62$ & $0.41$ & $0.42$ & $12$ & $50$ \\
GMM & $0.40$ & $0.66$ & $0.62$ & $0.41$ & $0.43$ & $62$ & $50$ \\
vMF-MM & $0.37$ & $0.65$ & $0.61$ & $0.38$ & $0.40$ & $267$ & $60$ \\
HAC & $0.42$ & $0.68$ & $0.65$ & $0.44$ & $0.45$ & $4$ & $55$ \\
CDP & $0.24$ & $0.68$ & $0.46$ & $0.26$ & $0.36$ & $2$ & $1,686$ \\
Spectral & $\boldsymbol{0.47}$ & $0.70$ & $\boldsymbol{0.68}$ & $\boldsymbol{0.48}$ & $\boldsymbol{0.50}$ & $99$ & $50$ \\
GCN-VE & $0.20$ & $0.65$ & $0.49$ & $0.22$ & $0.37$ & $1140$, $16$ & $905$ \\
STAR-FC & $0.36$ & $\boldsymbol{0.71}$ & $0.53$ & $0.37$ & $0.40$ & $341$, $1$ & $1,624$ \\
\bottomrule
\end{tabular}}
\end{center}

\begin{center}
\scalebox{0.96}{
\begin{tabular}{@{}cccccccc@{}}
\toprule
\multicolumn{8}{c}{Cars196} \\
\midrule
& Adj. Rand Index & NMI & AMI & $F_P$ & $F_B$ & Time to Cluster (m) & Number of Clusters \\
\midrule
DBSCAN & $0.29$ & $0.93$ & $0.49$ & $0.29$ & $0.58$ & $0.8$ & $17,801$ \\
MeanShift & $0.35$ & $0.93$ & $0.48$ & $0.35$ & $0.57$ & $54$ & $18,096$ \\
ARO & $0.52$ & $\boldsymbol{0.94}$ & $0.64$ & $0.52$ & $0.66$ & $0.2$ & $9,915$ \\
DP $k$-Means & $0.45$ & $0.93$ & $0.59$ & $0.45$ & $0.63$ & $13$ & $12,261$ \\
DP vMF $k$-Means & $0.45$ & $0.93$ & $0.59$ & $0.45$ & $0.63$ & $3$ & $11,282$ \\
$k$-Means & $0.46$ & $0.92$ & $0.58$ & $0.46$ & $0.61$ & $210$ & $7,750$ \\
Spherical $k$-Means & $0.45$ & $0.93$ & $0.59$ & $0.45$ & $0.62$ & $242$ & $8,250$ \\
HAC & $0.52$ & $0.93$ & $0.64$ & $0.52$ & $0.65$ & $4$ & $7,250$ \\
CDP & $\boldsymbol{0.56}$ & $\boldsymbol{0.94}$ & $\boldsymbol{0.67}$ & $\boldsymbol{0.56}$ & $\boldsymbol{0.69}$ & $0.1$ & $11,480$ \\
GCN-VE & $0.47$ & $0.93$ & $0.62$ & $0.47$ & $0.65$ & $20$, $13$s & $8,808$ \\
STAR-FC & $0.50$ & $0.93$ & $0.61$ & $0.50$ & $0.65$ & $8$, $2$s & $10,393$ \\
\bottomrule
\end{tabular}}
\end{center}

\caption{Clustering results for Cars196 and SOP. NMI, AMI, $F_P$, and $F_B$ represent normalized mutual information, adjusted mutual information, Pairwise F-score, and BCubed F-score, respectively. The time to cluster for Cars196 and SOP are measured in seconds and minutes, respectively. The time to cluster for GCN-VE and STAR-FC contains the train time and test time separated by a comma. Boldface indicates the highest value for a metric.}
\end{table}

\section{Comparison of $k$-Means Initialization Strategies}
\label{sec:k_means_initializations}

There are two common strategies for initializing clusters in $k$-means and spherical $k$-means.
The simplest is to initialize clusters by randomly selecting embeddings from the dataset.
The alternative is to use $k$-means++ \cite{Arthur2007} which tries to pick clusters that are generally distant from one another, leading to faster convergence and better expected performance.
We find that $k$-means++ indeed is a better initialization strategy, however, for large $k$ it increases the runtime drastically.
The increased runtime is caused by successively computing the nearest cluster-center for each embedding, and thus scales poorly as $k$ grows.
For our experiments, we use the $k$-means++ implementation from scikit-learn and admit that performance could be improved by running nearest-neighbor computation on accelerated hardware or by using approximate nearest-neighbor methods.

Based on our results summarized in the tables below, we find that the benefits of $k$-means++ initialization are not always significant (\eg, \cref{tab:k_means_dataset3}) and may not be worth the increased runtime, which can be upwards of 900x slower (e.g., SOP results in \cref{tab:k_means_cars_sop}) than a random initialization for large $k$.

\begin{table}[H]
\begin{center}
\begin{tabular}{@{}ccccc@{}}
\toprule
\multicolumn{5}{c}{Cars196} \\
\midrule
& Initialization & $F_P$ & $F_B$ & Time to Cluster (s) \\
\midrule
$k$-Means & Random & $\boldsymbol{0.41}$ & $0.42$ & $1$ \\
Spherical $k$-Means & Random & $\boldsymbol{0.41}$ & $\boldsymbol{0.43}$ & $1$ \\
$k$-Means & $k$-Means++ & $\boldsymbol{0.41}$ & $0.42$ & $12$ \\
Spherical $k$-Means & $k$-Means++ & $\boldsymbol{0.41}$ & $0.42$ & $12$ \\
\bottomrule
\end{tabular}
\end{center}

\begin{center}
\begin{tabular}{@{}ccccc@{}}
\toprule
\multicolumn{5}{c}{SOP} \\
\midrule
& Initialization & $F_P$ & $F_B$ & Time to Cluster (s) \\
\midrule
$k$-Means & Random & $0.36$ & $0.53$ & $14$ \\
Spherical $k$-Means & Random & $0.36$ & $0.53$ & $31$ \\
$k$-Means & $k$-Means++ & $\boldsymbol{0.46}$ & $0.61$ & $12,591$ \\
Spherical $k$-Means & $k$-Means++ & $0.45$ & $\boldsymbol{0.62}$ & $14,491$ \\
\bottomrule
\end{tabular}
\end{center}

\caption{Comparison of $k$-means initialization strategies on Cars196 and SOP. $F_P$ and $F_B$ represent Pairwise F-score and BCubed F-score, respectively. Boldface indicates the highest value for a metric.}
\label{tab:k_means_cars_sop}
\end{table}

\begin{table}[H]
\begin{center}
\begin{tabular}{@{}cccccccccccc@{}}
\toprule
& Initialization & \multicolumn{2}{c}{Test \#1} & \multicolumn{2}{c}{Test \#2} & \multicolumn{2}{c}{Test \#3} & \multicolumn{2}{c}{Test \#4} & \multicolumn{2}{c}{Test \#5} \\
\midrule
& & $F_P$ & $F_B$ & $F_P$ & $F_B$ & $F_P$ & $F_B$ & $F_P$ & $F_B$ & $F_P$ & $F_B$ \\
\midrule
Minibatch $k$-Means & Random & $\boldsymbol{0.64}$ & $0.65$ & $\boldsymbol{0.60}$ & $\boldsymbol{0.61}$ & $\boldsymbol{0.57}$ & $\boldsymbol{0.58}$ & $\boldsymbol{0.55}$ & $\boldsymbol{0.56}$ & $0.53$ & $0.54$ \\
Minibatch $k$-Means & $k$-Means++ & $\boldsymbol{0.64}$ & $\boldsymbol{0.66}$ & $\boldsymbol{0.60}$ & $\boldsymbol{0.61}$ & $\boldsymbol{0.57}$ & $\boldsymbol{0.58}$ & $\boldsymbol{0.55}$ & $\boldsymbol{0.56}$ & $\boldsymbol{0.54}$ & $\boldsymbol{0.55}$ \\
\bottomrule
\end{tabular}
\end{center}

\begin{center}
\begin{tabular}{@{}ccccccc@{}}
\toprule
& Initialization & Test \#1 & Test \#2 & Test \#3 & Test \#4 & Test \#5 \\
\midrule
Minibatch $k$-Means & Random  & $3$m & $7$m & $12$m & $18$m & $23$m \\
Minibatch $k$-Means & $k$-Means++ & $9$m & $41$m & $78$m & $147$m & $233$m \\
\bottomrule
\end{tabular}
\end{center}

\caption{Comparison of $k$-means initialization strategies on Dataset 3. For each of the five Dataset 3 test splits, the top table contains Pairwise ($F_P$) and BCubed ($F_B$) F-scores, and the bottom table contains the time to cluster in minutes. Boldface indicates the highest value for a metric.}
\label{tab:k_means_dataset3}
\end{table}

\section{Dataset 3 Degradation Study}
\label{sec:dataset3_degradation}

The table below contains clustering results for GCN-VE and HAC on test split \#1 of Dataset 3 where we reduce the dimensionality of the input embedding.
We reduce the dimensionality as a method for degrading the embedding discriminability, measured via Recall@1 in \cref{tab:dataset3_degradation_recall}.
These results are used to compute the harmonic means presented in \cref{fig:dataset3_degradation}.

\begin{table}[H]
\begin{center}
\begin{tabular}{@{}ccccccccc@{}}
\toprule
\multicolumn{9}{c}{Dataset 3 Test Split \#1} \\
\midrule
& \multicolumn{2}{c}{\multirow{2}{*}{32D}} & \multicolumn{2}{c}{\multirow{2}{*}{64D}} & \multicolumn{2}{c}{\multirow{2}{*}{128D}} & \multicolumn{2}{c}{256D} \\
& & & & & & & \multicolumn{2}{c}{(original)} \\
\midrule
& $F_P$ & $F_B$ & $F_P$ & $F_B$ & $F_P$ & $F_B$ & $F_P$ & $F_B$ \\
\midrule
HAC & $\boldsymbol{0.27}$ & $0.27$ & $\boldsymbol{0.53}$ & $0.54$ & $\boldsymbol{0.68}$ & $\boldsymbol{0.70}$ & $0.74$ & $0.75$ \\
GCN-VE & $0.26$ & $\boldsymbol{0.29}$ & $\boldsymbol{0.53}$ & $\boldsymbol{0.55}$ & $\boldsymbol{0.68}$ & $0.69$ & $\boldsymbol{0.78}$ & $\boldsymbol{0.79}$ \\
\bottomrule
\end{tabular}
\end{center}
\caption{Pairwise ($F_P$) and BCubed ($F_B$) F-scores for HAC and GCN-VE on test split \#1 of Dataset 3 as the embedding dimensionality increases. Boldface indicates the highest value for a metric.}
\label{tab:dataset3_degradation}
\end{table}

\section{Experimental Details}
\label{sec:experimental_details}

\subsection{Backbone}
\label{sec:backbone_details}

For Cars196 and SOP, the experimental details mimic Scott \etal \cite{Scott2021} unless explicitly noted.
For completeness, we recount the details below.

\subsubsection{Architecture}
\label{sec:backbone_architecture}

The backbone network architecture is a ResNet50 \cite{He2016}.
The network is initialized using weights pretrained on ImageNet and all batch-normalization parameters are frozen.
We remove the head of the architecture and add two fully-connected layers, with no activations, directly following the global-average-pooling layer.
The first fully-connected layer maps from 2048 units to 256 units, and the second fully-connected layer maps from 256 units to $Y$ units, where $Y$ is the number of classes in the backbone-training split of the dataset.
The embedding dimensionality is thus 256D.
For Cars196 and SOP, $Y = 49$ and $Y = 5658$, respectively.

\subsubsection{Dataset Augmentation}
\label{sec:dataset_augmentation}

\paragraph{Cars196.} Data augmentation during training includes: (1) resizing images to 256$\times$256, (2) random jittering of brightness, contrast, saturation, and hue with factors in [0.7, 1.3], [0.7, 1.3], [0.7, 1.3], and [-0.1, 0.1], respectively, (3) cropping at a random location with random size between 16\% and 100\% of the input size, (4) resizing the final cropped images to 224$\times$224, and (5) random horizontal flipping and $z$-score normalization.
Data augmentation during validation and testing includes: (1) resizing images to 256$\times$256, (2) center cropping to 224$\times$224, and (3) $z$-score normalization.
The mean and standard deviation for $z$-score normalization are the same values used for ImageNet.
We match Boudiaf \etal \cite{Boudiaf2020} and sample batches randomly.

\paragraph{SOP.} Data augmentation during training includes: (1) resizing images to 256$\times$256, (2) cropping at a random location with random size between 16\% and 100\% of the input size with the aspect ratio randomly selected in [0.75, 1.33], (3) resizing the final cropped images to 224$\times$224, and (4) random horizontal flipping and $z$-score normalization.
Data augmentation during validation and testing includes: (1) resizing images to 256$\times$256, (2) center cropping to 224$\times$224, and (3) $z$-score normalization.
The mean and standard deviation for $z$-score normalization are the same values used for ImageNet.
We match Boudiaf \etal \cite{Boudiaf2020} and sample batches randomly.

\subsubsection{Backbone Network Hyperparameters}
\label{sec:backbone_hparams}

Models are trained with SGD and either standard or Nesterov momentum.
Based on results from Scott \etal \cite{Scott2021}, we use cosine softmax cross-entropy as the loss.
For Cars196 and SOP, the validation split's Recall@1 is monitored throughout training and if it does not improve for 15 epochs, the learning rate is cut in half.
If the validation split's Recall@1 does not improve for 35 epochs, model training terminates early.
The model parameters are saved for the epoch resulting in the highest validation Recall@1.

The inverse-temperature, $\beta$, is parameterized as $\beta = \exp(\tau)$ where $\tau \in \mathbb{R}$ is an unconstrained network parameter learned automatically via gradient descent.
Unlike Scott \etal \cite{Scott2021}, $\tau$ is optimized using the same learning rate as all other network parameters.

The remaining hyperparameters for training the backbone network are:
\begin{itemize}
    \itemsep0.1em 
    \item Learning rate
    \item $\ell_2$ weight decay
    \item Momentum
    \item Nesterov momentum
    \item Initial value of $\tau$
\end{itemize}

Hyperparameters were optimized with a grid search. The table below contains the hyperparameter values used for the model achieving the best validation Recall@1.

\begin{table}[H]
\begin{center}
\begin{tabular}{@{}cccccc@{}}
\toprule
& Learning rate & $\ell_2$ weight decay & Momentum & Nesterov momentum & Initial value of $\tau$ \\
\midrule
Cars196 & $0.005$ & $5\times10^{-5}$ & $0.0$ & False & $0.0$ \\
SOP & $0.005$ & $5\times10^{-5}$ & $0.9$ & True & $-2.773$ \\
\bottomrule
\end{tabular}
\end{center}
\caption{Backbone hyperparameter values for Cars196 and SOP.}
\end{table}

\subsection{Clustering}
\label{sec:clustering_details}

For clustering, the inputs are the 256D embeddings from the penultimate output of the trained backbone network.
The embeddings are $\ell_2$-normalized unless explicitly noted otherwise.
To produce the embeddings for clustering, the images were augmented using the testing augmentation, which is: (1) resizing images to 256$\times$256, (2) center cropping to 224$\times$224, and (3) $z$-score normalization.

\subsubsection{GCN-VE Details}
\label{sec:gcn_ve_details}

We match Yang \etal \cite{Yang2020} and use a one-layer graph convolutional network (GCN) \cite{Kipf2017} with mean aggregation \cite{Yang2019} and ReLU activation for GCN-V, and a similar four-layer graph convolutional network for GCN-E.
GCN-V maps the 512D output of the GCN through a fully-connected layer to 512 units, then through a PReLU activation, followed by a final fully-connected layer to a single output unit.
The model is trained with mean-squared-error to match an empirically-estimated confidence based on the density of an embedding's neighborhood.
GCN-E has identical structure to GCN-V except that it contains a four-layer GCN, the GCN output is 256D, and the first fully-connected layer has 256 units.
GCN-E is trained with softmax cross-entropy to predict the likelihood that a pair of nearby embeddings belong to the same cluster.
The GCN layers are initialized using Xavier-uniform for the weights and zero for the biases.
The fully-connected layers are initialized using Kaiming-uniform for both weights and biases.

Both GCN-V and GCN-E are trained with SGD and standard momentum.
If the validation split's loss does not decrease for 100 epochs, the learning rate is cut in half, and if it does not decrease for 250 epochs, model training terminates early.
The model parameters are saved for the epoch resulting in the lowest validation loss.

\subsubsection{STAR-FC Details}
\label{sec:star_fc_details}

We match Shen \etal \cite{Shen2021} and use a one-layer graph convolutional network (GCN) \cite{Kipf2017} with mean aggregation \cite{Yang2019} and ReLU activation.
The GCN output is 1024D which is then passed through a fully-connected layer to 512 units, then through a PReLU activation, followed by a final fully-connected layer to a single output unit.
The model is trained with softmax cross-entropy to predict the likelihood that a pair of embeddings belong to the same cluster.
The GCN layers are initialized using Xavier-uniform for the weights and zero for the biases.
The fully-connected layers are initialized using Kaiming-uniform for both weights and biases.

STAR-FC is trained with SGD and standard momentum.
If the validation split's loss does not decrease for 15 epochs, the learning rate is cut in half, and if it does not decrease for 35 epochs, model training terminates early.
The model parameters are saved for the epoch resulting in the lowest validation loss.

\subsubsection{Clustering Hyperparameters}
\label{sec:clustering_hparams}

For each clustering method, we list the hyperparameters and their values for Cars196 and SOP.
All hyperparameters were optimized with a grid search.

\paragraph{$k$-Means with Random Initialization \cite{Lloyd1982,Sculley2010}.} The only hyperparameters are $k$, the number of clusters, and the minibatch size.
A minibatch size of -1 indicates the batch size is equal to the full dataset size.

\begin{table}[H]
\begin{center}
\begin{tabular}{@{}ccc@{}}
\toprule
& $k$ & Minibatch size \\
\midrule
Cars196 & $40$ & $-1$ \\
SOP & $8,250$ & $-1$ \\
\bottomrule
\end{tabular}
\end{center}
\caption{Hyperparameter values for $k$-means with random initialization.}
\end{table}

\paragraph{$k$-Means with $k$-Means++ Initialization \cite{Arthur2007,Sculley2010}.} The only hyperparameters are $k$, the number of clusters, and the minibatch size.
A minibatch size of -1 indicates the batch size is equal to the full dataset size.

\begin{table}[H]
\begin{center}
\begin{tabular}{@{}ccc@{}}
\toprule
& $k$ & Minibatch size \\
\midrule
Cars196 & $50$ & $-1$ \\
SOP & $7,750$ & $-1$ \\
\bottomrule
\end{tabular}
\end{center}
\caption{Hyperparameter values for $k$-means with $k$-means++ initialization.}
\end{table}

\paragraph{Spherical $k$-Means with Random Initialization \cite{Hornik2012}.} The only hyperparameter is $k$, the number of clusters.
Due to the performance of spherical $k$-means closely matching that of $k$-means with $\ell_2$-normalized embeddings, and the lack of a readily-available implementation of minibatch spherical $k$-means, we only ran spherical $k$-means for Cars196 and SOP, where the minibatch size could be -1.

\begin{table}[H]
\begin{center}
\begin{tabular}{@{}cc@{}}
\toprule
& $k$ \\
\midrule
Cars196 & $45$  \\
SOP & $8,000$ \\
\bottomrule
\end{tabular}
\end{center}
\caption{Hyperparameter values for spherical $k$-means with random initialization.}
\end{table}

\paragraph{Spherical $k$-Means with $k$-Means++ Initialization \cite{Hornik2012}.} The only hyperparameter is $k$, the number of clusters.
Due to the performance of spherical $k$-means closely matching that of $k$-means with $\ell_2$-normalized embeddings, and the lack of a readily-available implementation of minibatch spherical $k$-means, we only ran spherical $k$-means for Cars196 and SOP, where the minibatch size could be -1.

\begin{table}[H]
\begin{center}
\begin{tabular}{@{}cc@{}}
\toprule
& $k$ \\
\midrule
Cars196 & $50$ \\
SOP & $8,250$ \\
\bottomrule
\end{tabular}
\end{center}
\caption{Hyperparameter values for spherical $k$-means with $k$-means++ initialization.}
\end{table}

\paragraph{Dirichlet-Process (DP) $k$-Means \cite{Kulis2012}.} The hyperparameters are:
\begin{itemize}
    \itemsep0.1em 
    \item $\lambda$, the cluster penalty
    \item Initialization strategy, either `Global Centroid' or `Random'
    \item Whether to do an online EM update
\end{itemize}

\begin{table}[H]
\begin{center}
\begin{tabular}{@{}cccc@{}}
\toprule
& $\lambda$ & Initialization strategy & Online EM \\
\midrule
Cars196 & $0.9$ & Global Centroid & True \\
SOP & $0.85$ & Random & False \\
\bottomrule
\end{tabular}
\end{center}
\caption{Hyperparameter values for DP $k$-means.}
\end{table}

\paragraph{Dirichlet-Process (DP) von Mises--Fisher (vMF) $k$-Means \cite{Straub2015}.} The hyperparameters are:
\begin{itemize}
    \itemsep0.1em 
    \item $\lambda$, the cluster penalty
    \item Initialization strategy, either `Global Centroid' or `Random'
    \item Whether to do an online EM update
\end{itemize}

\begin{table}[H]
\begin{center}
\begin{tabular}{@{}cccc@{}}
\toprule
& $\lambda$ & Initialization strategy & Online EM \\
\midrule
Cars196 & $-0.55$ & Random & True \\
SOP & $-0.38$ & Random & False \\
\bottomrule
\end{tabular}
\end{center}
\caption{Hyperparameter values for DP vMF $k$-means.}
\end{table}

\paragraph{Gaussian Mixture Model (GMM).} The hyperparameters are:
\begin{itemize}
    \itemsep0.1em 
    \item $k$, the number of clusters
    \item Initialization strategy, either `k-Means' or `Random'
    \item Type of covariance, either `Full' or `Diagonal'
\end{itemize}
GMM was run only on Cars196 due to runtime inefficiencies.

\begin{table}[H]
\begin{center}
\begin{tabular}{@{}cccc@{}}
\toprule
& $k$ & Initialization strategy & Covariance type \\
\midrule
Cars196 & $50$ & $k$-Means & Diagonal \\
\bottomrule
\end{tabular}
\end{center}
\caption{Hyperparameter values for GMM.}
\end{table}

\paragraph{von Mises--Fisher Mixture Model (vMF-MM) \cite{Gopal2014,Banerjee2005}.} The hyperparameters are:
\begin{itemize}
    \itemsep0.1em 
    \item $k$, the number of clusters
    \item Initialization strategy, one of `k-Means++' or `Random', `Random-Class', or `Random-Orthonormal'
    \item Posterior type, either `Hard' or `Soft'
\end{itemize}
vMF-MM was run only on Cars196 due to runtime inefficiencies.

\begin{table}[H]
\begin{center}
\begin{tabular}{@{}cccc@{}}
\toprule
& $k$ & Initialization strategy & Posterior type \\
\midrule
Cars196 & $50$ & Random-Class & Soft \\
\bottomrule
\end{tabular}
\end{center}
\caption{Hyperparameter values for vMF-MM.}
\end{table}

\paragraph{Hierarchical Agglomerative Clustering (HAC) \cite{Sibson1973}.} The hyperparameters are:
\begin{itemize}
    \itemsep0.1em 
    \item $k$, the number of clusters
    \item Affinity metric, one of `Euclidean' or `Cosine'
    \item Linkage criterion, one of `Ward', `Complete', `Average', or `Single'
    \item Number of neighbors for computing the nearest-neighbor affinity subgraph
\end{itemize}

\begin{table}[H]
\begin{center}
\begin{tabular}{@{}ccccc@{}}
\toprule
& $k$ & Affinity metric & Linkage criterion & Number of neighbors \\
\midrule
Cars196 & $55$ & Euclidean & Ward & $40$ \\
SOP & $7,250$ & Euclidean & Ward & $2$ \\
\bottomrule
\end{tabular}
\end{center}
\caption{Hyperparameter values for HAC.}
\end{table}

\paragraph{Approximate Rank Order (ARO) \cite{Otto2018}.} The hyperparameters are:
\begin{itemize}
    \itemsep0.1em 
    \item Number of neighbors for computing the nearest-neighbors subgraph
    \item Threshold on rank-order distances
    \item Distance metric, one of `Euclidean' or `Cosine'
\end{itemize}
Note that for $\ell_2$-normalized embeddings, both Euclidean and cosine distance metrics produce identical results because the rank-order is unchanged.

\begin{table}[H]
\begin{center}
\begin{tabular}{@{}cccc@{}}
\toprule
& Number of neighbors & Threshold & Distance metric \\
\midrule
Cars196 & $30$ & $0.65$ & Euclidean \\
SOP & $5$ & $1.0$ & Euclidean \\
\bottomrule
\end{tabular}
\end{center}
\caption{Hyperparameter values for ARO.}
\end{table}

\paragraph{Density-Based Spatial Clustering of Applications with Noise (DBSCAN) \cite{Ester1996}.} The hyperparameters are:
\begin{itemize}
    \itemsep0.1em 
    \item $\epsilon$, the maximum distance between two embeddings in the same neighborhood
    \item Minimum number of samples in neighborhood for core point
    \item Distance metric, either `Euclidean' or `Cosine'
    \item Whether a sparse matrix is used for nearest-neighbor subgraph
\end{itemize}

\begin{table}[H]
\begin{center}
\begin{tabular}{@{}ccccc@{}}
\toprule
& $\epsilon$ & Minimum number of samples & Distance metric & Sparse matrix \\
\midrule
Cars196 & $0.25$ & $5$ & Cosine & False \\
SOP & $0.66$ & $2$ & Euclidean & False \\
\bottomrule
\end{tabular}
\end{center}
\caption{Hyperparameter values for DBSCAN.}
\end{table}

\paragraph{MeanShift \cite{Cheng1995}.} The hyperparameters are:
\begin{itemize}
    \itemsep0.1em 
    \item Bandwidth for RBF kernel
    \item Minimum bin frequency
    \item Whether to cluster all embeddings including orphans
\end{itemize}

\begin{table}[H]
\begin{center}
\begin{tabular}{@{}cccc@{}}
\toprule
& Bandwidth & Minimum bin frequency & Cluster all embeddings \\
\midrule
Cars196 & $0.75$ & $5$ & True \\
SOP & $0.65$ & 1 & False \\
\bottomrule
\end{tabular}
\end{center}
\caption{Hyperparameter values for MeanShift.}
\end{table}

\paragraph{Spectral Clustering \cite{Ng2001}.} The hyperparameters are:
\begin{itemize}
    \itemsep0.1em 
    \item $k$, the number of clusters
    \item Method for constructing the affinity matrix
    \item Number of neighbors for computing the nearest-neighbor affinity subgraph
    \item Number of components for the spectral embedding
\end{itemize}
Spectral clustering was run only on Cars196 due to runtime inefficiencies.

\begin{table}[H]
\begin{center}
\begin{tabular}{@{}ccccc@{}}
\toprule
& $k$ & Affinity & Number of neighbors & Number of components \\
\midrule
Cars196 & $50$ & Nearest neighbor subgraph & $20$ & $50$ \\
\bottomrule
\end{tabular}
\end{center}
\caption{Hyperparameter values for spectral clustering.}
\end{table}

\paragraph{Consensus-Driven Propagation (CDP) \cite{Zhan2018}.} The hyperparameters are:
\begin{itemize}
    \itemsep0.1em 
    \item Number of neighbors for computing the nearest-neighbor subgraph
    \item Threshold on cosine similarity in the nearest-neighbor subgraph for pruning edges
    \item Threshold step size
    \item Max cluster size
\end{itemize}

\begin{table}[H]
\begin{center}
\begin{tabular}{@{}ccccc@{}}
\toprule
& Number of neighbors & Threshold & Threshold step size & Max cluster size \\
\midrule
Cars196 & $2$ & $0.6$ & $0.01$ & $320$ \\
SOP & $5$ & $-0.2$ & $0.05$ & $10$ \\
\bottomrule
\end{tabular}
\end{center}
\caption{Hyperparameter values for CDP.}
\end{table}

\paragraph{Tree Deduction \cite{Yang2020}.} The hyperparameters are the number of neighbors for computing the nearest-neighbor subgraph and a threshold on cosine similarity for pruning edges during tree deduction.

\begin{table}[H]
\begin{center}
\begin{tabular}{@{}ccc@{}}
\toprule
& Number of neighbors & Threshold \\
\midrule
Cars196 & $2$ & $0.75$ \\
SOP & $1$ & $0.6$ \\
\bottomrule
\end{tabular}
\end{center}
\caption{Hyperparameter values for tree deduction.}
\end{table}

\paragraph{Tree Deduction with Embedding $\ell_2$-Norm Confidence.} The hyperparameters are the number of neighbors for computing the nearest-neighbor subgraph and a threshold on cosine similarity for pruning edges during tree deduction.

\begin{table}[H]
\begin{center}
\begin{tabular}{@{}ccc@{}}
\toprule
& Number of neighbors & Threshold \\
\midrule
Cars196 & $20$ & $0.6$ \\
SOP & $3$ & $0.65$ \\
\bottomrule
\end{tabular}
\end{center}
\caption{Hyperparameter values for tree deduction with embedding $\ell_2$-norm confidence.}
\end{table}

\paragraph{Embedding $\ell_2$-Norm Confidence + GCN-E + Tree Deduction.} The hyperparameters are:
\begin{itemize}
    \itemsep0.1em 
    \item Number of neighbors for computing the adjacency matrix (N)
    \item Threshold on cosine similarity for pruning edges in the adjacency matrix ($\tau_1$)
    \item Ignore ratio (IR)
    \item Threshold on cosine similarity for pruning edges during tree deduction ($\tau_2$)
    \item Number of units in the hidden layers of the network (H)
    \item Learning rate (LR)
    \item Momentum (MOM)
    \item $\ell_2$ weight decay ($\ell_2$)
    \item Dropout probability (D)
\end{itemize}

\begin{table}[H]
\begin{center}
\begin{tabular}{@{}cccccccccc@{}}
\toprule
& N & $\tau_1$ & IR & $\tau_2$ & H & LR & MOM & $\ell_2$ & D \\
\midrule
Cars196 & $60$ & $0.0$ & $0.1$ & $0.6$ & $512$ & $0.01$ & $0.9$ & $1\times10^{-5}$ & $0.2$ \\
SOP & $5$ & $0.0$ & $0.7$ & $0.7$ & $512$ & $0.01$ & $0.9$ & $1\times10^{-5}$ & $0.0$ \\
\bottomrule
\end{tabular}
\end{center}
\caption{Hyperparameter values for embedding $\ell_2$-norm confidence + GCN-E + tree deduction.}
\end{table}

\paragraph{GCN-VE \cite{Yang2020}.} GCN-VE has two sub-networks, GCN-V and GCN-E.
We present the hyperparameters associated with each sub-network independently.

The hyperparameters for GCN-V are:
\begin{itemize}
    \itemsep0.1em 
    \item Number of neighbors for computing the adjacency matrix (N)
    \item Threshold on cosine similarity for pruning edges in the adjacency matrix ($\tau$)
    \item Number of units in the hidden layers of the network (H)
    \item Learning rate (LR)
    \item Momentum (MOM)
    \item $\ell_2$ weight decay ($\ell_2$)
    \item Dropout probability (D)
\end{itemize}

\begin{table}[H]
\begin{center}
\begin{tabular}{@{}cccccccc@{}}
\toprule
& N & $\tau$ & H & LR & MOM & $\ell_2$ & D \\
\midrule
Cars196 & $10$ & $0.0$ & $512$ & $0.01$ & $0.9$ & $1\times10^{-5}$ & $0.0$ \\
SOP & $2$ & $0.0$ & $512$ & $0.1$ & $0.9$ & $1\times10^{-5}$ & $0.0$ \\
\bottomrule
\end{tabular}
\end{center}
\caption{Hyperparameter values for GCN-E.}
\end{table}

The hyperparameters for GCN-E are:
\begin{itemize}
    \itemsep0.1em 
    \item Number of neighbors for computing the adjacency matrix (N)
    \item Threshold on cosine similarity for pruning edges in the adjacency matrix ($\tau_1$)
    \item Ignore ratio (IR)
    \item Threshold on cosine similarity for pruning edges during tree deduction ($\tau_2$)
    \item Number of units in the hidden layers of the network (H)
    \item Learning rate (LR)
    \item Momentum (MOM)
    \item $\ell_2$ weight decay ($\ell_2$)
    \item Dropout probability (D)
\end{itemize}

\begin{table}[H]
\begin{center}
\begin{tabular}{@{}cccccccccc@{}}
\toprule
& N & $\tau_1$ & IR & $\tau_2$ & H & LR & MOM & $\ell_2$ & D \\
\midrule
Cars196 & $200$ & $0.0$ & $0.0$ & $0.7$ & $512$ & $0.01$ & $0.9$ & $1\times10^{-5}$ & $0.0$ \\
SOP & $5$ & $0.0$ & $0.7$ & $0.8$ & $512$ & $0.01$ & $0.9$ & $1\times10^{-5}$ & $0.0$ \\
\bottomrule
\end{tabular}
\end{center}
\caption{Hyperparameter values for GCN-E.}
\end{table}

\paragraph{STAR-FC \cite{Shen2021}.} The hyperparameters are:
\begin{itemize}
    \itemsep0.1em 
    \item Number of neighbors for computing the adjacency matrix (N)
    \item Threshold on cosine similarity for pruning edges in the adjacency matrix ($\tau$)
    \item Number of seed clusters (SC)
    \item Number of nearest clusters (NC)
    \item Number of random clusters (RC)
    \item Random node proportion (RNP)
    \item Prune threshold (P)
    \item Intimacy threshold (I)
    \item Number of units in the hidden layers of the network (H)
    \item Learning rate (LR)
    \item Momentum (MOM)
    \item $\ell_2$ weight decay ($\ell_2$)
\end{itemize}

\begin{table}[H]
\begin{center}
\begin{tabular}{@{}cccccccccccccc@{}}
\toprule
& N & $\tau$ & SC & NC & RC & RNP & P & I & H & LR & MOM & $\ell_2$ \\
\midrule
Cars196 & $10$ & $0.0$ & $4$ & $6$ & $10$ & $0.9$ & $0.15$ & $0.6$ & $512$ & $0.1$ & $0.9$ & $1\times10^{-5}$ \\
SOP & $3$ & $0.0$ & $4$ & $500$ & $250$ & $0.9$ & $0.5$ & $0.7$ & $512$ & $0.1$ & $0.9$ & $1\times10^{-5}$ \\
\bottomrule
\end{tabular}
\end{center}
\caption{Hyperparameter values for STAR-FC.}
\end{table}

}

\end{document}